%% file: main.tex
\documentclass[twoside,11pt]{article}

\usepackage[preprint]{jmlr2e}

\usepackage{amsmath}
\usepackage{amsfonts}
\usepackage{dsfont}
\usepackage{mathtools}
\usepackage{physics2}
\usephysicsmodule{ab,doubleprod}
\usepackage{cleveref}

\usepackage{listings}
\usepackage[dvipsnames]{xcolor}

\definecolor{ghblue}{rgb}{0.0, 0.33, 0.62}
\definecolor{ghpurple}{rgb}{0.42, 0.0, 0.55}
\definecolor{ghgreen}{rgb}{0.0, 0.5, 0.0}
\definecolor{ghred}{rgb}{0.82, 0.10, 0.14}
\definecolor{ghgray}{rgb}{0.38, 0.42, 0.47}
\definecolor{ghbackground}{rgb}{0.99, 0.99, 0.99}
\definecolor{ghborder}{rgb}{0.85, 0.86, 0.88}

\lstdefinestyle{githublightstyle}{
    backgroundcolor=\color{ghbackground},
    commentstyle=\color{ghgray}\itshape,
    keywordstyle=\color{ghred}\bfseries,
    keywordstyle=[2]\color{ghblue},
    keywordstyle=[3]\color{ghpurple},
    numberstyle=\tiny\color{ghgray},
    stringstyle=\color{ghgreen},
    basicstyle=\ttfamily\small,
    breakatwhitespace=false,
    breaklines=true,
    captionpos=b,
    keepspaces=true,
    numbers=left,
    numbersep=8pt,
    showspaces=false,
    showstringspaces=false,
    showtabs=false,
    tabsize=4,
    frame=single,
    framesep=3pt,
    rulecolor=\color{ghborder},
    xleftmargin=10pt,
    xrightmargin=10pt,
    morekeywords=[2]{self, None, True, False},
    morekeywords=[3]{torch, nn, Module, Parameter, tensor}
}

\usepackage{lastpage}
\jmlrheading{}{2026}{1-\pageref{LastPage}}{}{}{}{Thorsten Kurth, Max Rietmann, Mauro Bisson, Andrea Paris, Alberto Carpentieri, Jean Kossaifi, Anima Anandkumar, Christian Hundt and Boris Bonev}

\ShortHeadings{Differentiable Signal Processing on the Sphere}{Kurth et al.}
\firstpageno{1}

\begin{document}

\title{A library for differentiable signal processing and machine learning on the sphere}

\author{
    \begin{center}
       \name Thorsten Kurth$^{1,\dagger}$,
       \name Max Rietmann$^{1}$,
       \name Mauro Bisson$^{1}$, \\
       \name Andrea Paris$^{1}$,
       \name Alberto Carpentieri$^{1}$,
       \name Jean Kossaifi$^{1}$, \\
       \name Anima Anandkumar$^{1,2}$,
       \name Christian Hundt$^{1}$,
       \name Boris Bonev$^{1,\dagger}$ \\
       \addr $^{1}$NVIDIA Corporation \quad
       $^{2}$California Institute of Technology \quad
       $^{\dagger}$Equal contribution
    \end{center}
}

\editor{To be assigned}

\maketitle

\begin{abstract}
The two-dimensional sphere embedded in three-dimensional Euclidean space $S^2$, plays a central role in a variety of scientific and engineering domains, including geophysics, planetary science, geodesy, atmospheric physics, quantum chemistry, cosmology, and virtual reality, among many others. As machine learning increasingly permeates these fields, the demand grows for robust tools that process and model functions on the sphere, while respecting the inherent topological and symmetry properties of the domain. We present \texttt{torch-harmonics}, a comprehensive library that offers efficient, differentiable implementations of advanced signal processing and machine learning (ML) methods for spherical data. These include the spherical harmonic transform (SHT), the spherical analogue of the Fourier transform, vector spherical harmonics, discrete-continuous and spectral convolutions, as well as both global and neighborhood spherical attention mechanisms. Beyond traditional representations, \texttt{torch-harmonics} provides the building blocks for state-of-the-art spherical ML architectures such as spherical transformers in order to enable scalable, rotationally-aware learning and inference in modern scientific and engineering applications.
\end{abstract}


\input{sections/introduction}
\input{sections/functionality}
\input{sections/ml_models}
\input{sections/applications}
\input{sections/conclusion}

\newpage

\bibliography{references}

\newpage

\appendix
\input{sections/appendix/signal_processing}
\input{sections/appendix/code_examples}

\input{sections/appendix/performance_optimization}

\end{document}

%% file: sections/introduction.tex
\section{Introduction}
\label{sec:introduction}

The two-dimensional sphere $S^2$ plays a central role in many scientific domains, including geophysics, atmospheric physics, cosmology, and computer graphics. As machine learning becomes increasingly prevalent in these fields, there is growing demand for tools that process spherical signals while respecting the sphere's topological and symmetry properties.

Standard deep learning operations designed for Euclidean domains do not naturally extend to the sphere without introducing distortions or singularities. Geometric deep learning approaches such as graph neural networks on spherical meshes and equivariant networks based on spherical harmonics have been proposed, but their practical adoption has been hindered by the lack of efficient, differentiable, and scalable implementations \citep{Bronstein2021,Cohen2016,Cohen2018,Esteves2017,Esteves2020a,Esteves2023,Ocampo2022,Cobb2020,Defferrard2020,Brehmer2025}.

We present \texttt{torch-harmonics}, a PyTorch~\citep{Paszke2019} library for differentiable signal processing on the sphere. The library provides efficient implementations of spherical harmonic transforms (SHT), vector spherical harmonic transforms~\citep{Schaeffer2013}, discrete-continuous (DISCO) convolutions, spherical attention mechanisms and other important operations on the spherical domain. Through custom CUDA~\citep{Nickolls2008} kernels and distributed computing strategies, \texttt{torch-harmonics} enables training of high-resolution models that were previously computationally prohibitive. The library serves as the foundation for architectures such as the Spherical Fourier Neural Operator (SFNO) \citep{Bonev2023} and FourCastNet3 (FCN3) \citep{Bonev2025b}.

\begin{figure}[t]
    \centering
    \includegraphics[width=0.79\textwidth]{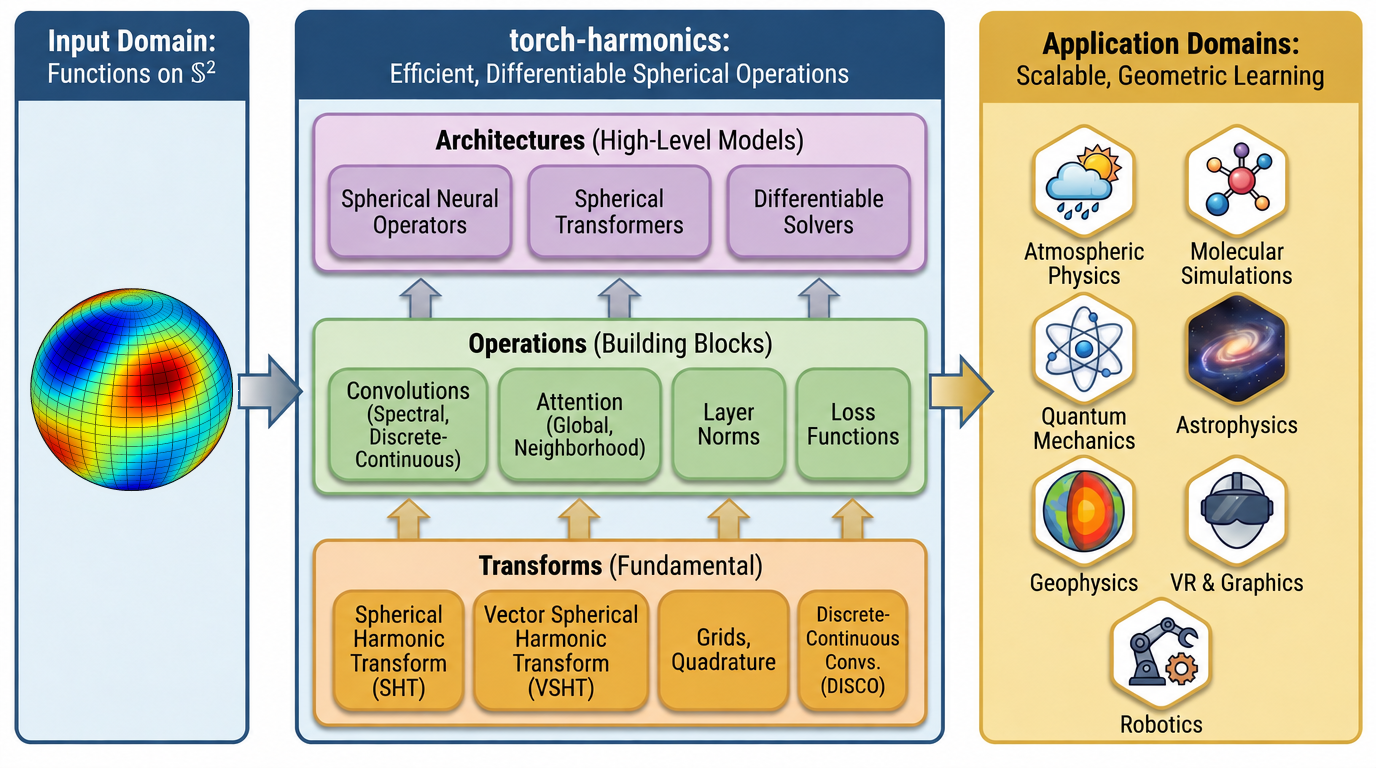}
    \caption{Overview of \texttt{torch-harmonics}: Spherical signal processing fundamentals enable high-level models such as spherical neural operators and differentiable PDE solvers. Applications include weather prediction, molecular simulations and more.}
    \label{fig:use_cases}
\end{figure}

%% file: sections/functionality.tex
\section{Library Design and Functionality}
\label{sec:functionality}

\texttt{torch-harmonics} provides efficient differentiable signal processing and machine learning operations for spherical data that integrate seamlessly into PyTorch models and training pipelines. The main design goals are: \textbf{Differentiability}---all operations support backpropagation via PyTorch's \texttt{autograd}; \textbf{Efficiency}---custom CUDA kernels for intensive operations; \textbf{Compatibility}---fallback implementations in pure PyTorch, and \textbf{Scalability}---distributed memory computing support to process high-resolution spherical data.

\subsection{Core Signal Processing Operations}

Three commonly used grid types are currently supported: equiangular/equirectangular (e.g. classical latitude-longitude grid used in geosciences), Legendre-Gauss grids and Gauss-Lobatto grids. Support for other popular grid types is planned.

\paragraph{Spherical Harmonic Transforms}
The Spherical Harmonic Transform (SHT) generalizes the Fourier transform on the sphere, decomposing values defined on the equiangular grid into a series of spherical harmonic basis functions $Y_\ell^m$, capturing the signal's energy at different spatial scales. \texttt{torch\-harmonics} provides forward and inverse transforms (\texttt{RealSHT} and \texttt{InverseRealSHT}). This computation can be decomposed into a Fourier transform and a Legendre transform, using efficient implementations of the fast Fourier transform (FFT) and batched matrix multiplications \citep{Schaeffer2013}.

\paragraph{Vector Spherical Harmonic Transforms}
\texttt{RealVectorSHT}, \texttt{InverseRealVectorSHT} decompose vector fields into divergence\-free
and curl-free components by applying the forward SHT to the potentials of the vector field. This enables efficient computation of differential operators such as divergence and curl directly in the spectral domain (see \Cref{app:signal_processing}), which is particularly useful for solving partial differential equations (PDEs) on the sphere.

\paragraph{Resampling and Interpolation}
The library supports resampling between different spherical grids using spectral interpolation, which enables alias\-free resolution changes. Bilinear interpolation in spherical coordinates based on Haversine distance is also supported.

\paragraph{Quadrature on the Sphere}
Quadrature rules for all supported grid types are essential for computing inner products, norms, and integrals while respecting the spherical geometry.

\paragraph{Spectral/Discrete-Continuous Convolutions}
Convolution on the sphere is defined via the action of the rotation group $SO(3)$.
Two complementary approaches are supported: Spectral Convolutions, and Discrete-Continuous (DISCO) Convolutions, which support local, anisotropic filters through direct quadrature of the rotated filter kernel.
\texttt{torch\-harmonics} offers a wide variety of filter basis functions for the DISCO convolution, such as piecewise linear hat functions, Zernike polynomials and wavelet-like functions.

\paragraph{Spherical Attention}
The library provides implementations of global and local (neighborhood) attention on the sphere. Quadrature weights ensure proper integration over spherical domains, respecting the non-uniform sampling density of spherical grids.

\paragraph{Angular Power Spectrum}
The angular power spectrum quantifies the distribution of energy across scales and provides an important diagnostic for verifying spectral signatures.

\paragraph{Distributed Memory Parallelism}
All these components also support parallel distributed memory computing via \texttt{torch.distributed}. This enhances computational performance and reduces memory footprint by distributing the spherical signal across ranks. The distributed SHT, VSHT and DISCO modules use a pencil decomposition strategy, while attention mechanisms use halo exchanges to provide distributed implementations.

%% file: sections/ml_models.tex
\section{Machine Learning Architectures}
\label{sec:ml_models}
\paragraph{Spherical Fourier Neural Operators} Spherical Fourier Neural Operators \citep{Bonev2023} generalize the Fourier neural operator \citep{Li2020b} to the sphere, implementing $SO(3)$ group convolutions via the spherical convolution theorem \citep{Driscoll1994}. SFNO blocks operate in frequency domain: inputs are transformed via SHT, multiplied with learnable filters in the spectral domain, and transformed back via inverse SHT, followed by pointwise nonlinearity. This provides global receptive fields while respecting spherical topology. SFNO architectures power several weather and climate models \citep{Bonev2023,Watt-Meyer2023,Watt-Meyer2025,Mahesh2024a,Mahesh2024b,Guan2025}.

\paragraph{Local Spherical Neural Operators}
Local Spherical Neural Operators \citep{Liu2024} employ DISCO convolutions with compact support to learn anisotropic~\citep{Ocampo2022}, spatially localized features while maintaining approximate rotation equivariance. These are well-suited for processes with preferential directions such as boundary layers or topographic effects. FourCastNet 3 (FCN3) \citep{Bonev2025b} combines SFNO and LSNO layers to capture both global wave dynamics and local atmospheric features.

\paragraph{Spherical Transformers}
Spherical Transformers \citep{Bonev2025a} generalize vision transformers to spherical geometry through continuous attention formulations. \texttt{torch\-harmonics} provides accelerated implementations of global and neighborhood attention mechanisms, using geodesic distances and proper quadrature weights for spherical grids. These are particularly useful for data assimilation and processing irregularly sampled observations \citep{Bonev2025a, Lang2024, Gupta2026}.

\paragraph{Hybrid Physics-ML Architectures}
The differentiable spectral transforms enable hybrid methods incorporating machine learning into traditional numerical solvers. The repository includes a differentiable spectral solver for the shallow water equations demonstrating this capability. Such solvers form the backbone of global circulation models and enable training of neural GCMs \citep{Kochkov2023}.

%% file: sections/applications.tex
\section{Applications and Examples}
\label{sec:applications}

\texttt{torch-harmonics} has been deployed in production weather forecasting systems and scientific computing applications \citep{Bonev2023,Watt-Meyer2023,Watt-Meyer2025,Mahesh2024a,Mahesh2024b,Guan2025,Bonev2025b,Mansouri2025,Leonardi2026}. Complete example implementations in the repository include: \textbf{Differentiable shallow water solver}---a spectral solver for the shallow water equations on the sphere using VSHT for divergence and curl computation, enabling gradient-based parameter estimation and differentiable GCM training (see \Cref{app:swe_solver}); \textbf{Spherical neural operators on PDE data}---training SFNO and LSNO models on shallow water equation data generated by the differentiable solver; \textbf{Spherical image processing}---depth estimation and segmentation on 360° images using Spherical Transformers, compared against Euclidean baselines; and \textbf{Code examples}---complete implementations of spherical layers, models and solvers, demonstrating usage and integration into standard training pipelines (\Cref{app:code_examples}).

%% file: sections/conclusion.tex
\section{Conclusion}
\label{sec:conclusion}

We presented \texttt{torch-harmonics}, an open-source library for differentiable signal processing and machine learning on the sphere.
The library provides efficient, GPU-accelerated implementations of spherical harmonic transforms, discrete-continuous convolutions, and spherical attention mechanisms, all fully integrated into the PyTorch ecosystem.
Custom CUDA kernels and distributed memory parallelism enable scaling to high resolutions that were previously computationally prohibitive.
Beyond weather and climate applications \citep{Bonev2023,Watt-Meyer2025}, \texttt{torch-harmonics} has enabled applications in solar wind modeling \citep{Mansouri2025}, omnidirectional 3D vision \citep{Bonev2025a, Wang2026}, and molecular force fields \citep{Leonardi2026}, underscoring its role as a general-purpose tool for learning and signal processing on the sphere.

%% file: sections/appendix/signal_processing.tex
\section{Signal processing on the sphere}
\label{app:signal_processing}

In this appendix, we provide the mathematical background for the signal processing operations implemented in \texttt{torch-harmonics}. We begin with the definition of the spherical domain and coordinate systems, followed by a discussion of functions on the sphere and the rotation group $SO(3)$. We then introduce the Spherical Harmonic Transform (SHT), spherical convolutions, and attention mechanisms.

\subsection{Coordinate systems and functions on the sphere}

The unit sphere $S^2$ is the set of points $x \in \mathbb{R}^3$ with unit norm $\|x\|_2 = 1$. We parameterize $S^2$ using spherical coordinates $(\vartheta, \varphi)$, where $\vartheta \in [0, \pi]$ denotes the colatitude (with $\vartheta=0$ at the North Pole) and $\varphi \in [0, 2\pi)$ denotes the longitude:
\begin{equation}
    x(\vartheta, \varphi) =
    \begin{bmatrix}
        \sin \vartheta \cos \varphi\\
        \sin \vartheta \sin \varphi\\
        \cos \vartheta
    \end{bmatrix}.
\end{equation}
We consider real-valued square-integrable functions $u: S^2 \rightarrow \mathbb{R}$ in the Hilbert space $L^2(S^2)$, equipped with the inner product
\begin{equation}
    \langle u, v \rangle_{L^2(S^2)} = \int_{S^2} u(x) v(x) \mathrm{d}\mu(x) = \int_{0}^{2\pi} \int_{0}^{\pi} u(\vartheta, \varphi) v(\vartheta, \varphi) \sin \vartheta \, \mathrm{d}\vartheta \, \mathrm{d}\varphi,
\end{equation}
where $\mathrm{d}\mu(x) = \sin \vartheta \, \mathrm{d}\vartheta \, \mathrm{d}\varphi$ is the standard rotation-invariant Lebesgue measure on the sphere.

\subsection{The rotation group SO(3)}
The Special Orthogonal group $SO(3)$ consists of all $3 \times 3$ orthogonal matrices with determinant $+1$. Elements $R \in SO(3)$ represent rotations in three-dimensional space. $SO(3)$ acts transitively on $S^2$ via matrix-vector multiplication $x \mapsto Rx$.
A function $u \in L^2(S^2)$ can be rotated by an operator $\mathcal{R}$ associated with $R \in SO(3)$ as:
\begin{equation}
    [\mathcal{R} u](x) = u(R^{-1}x).
\end{equation}
This action preserves the inner product, i.e., $\langle \mathcal{R} u, \mathcal{R} v \rangle = \langle u, v \rangle$.

\subsection{Grids and Quadrature}
\label{app:grids_quad}
Spherical signals $u: S^2 \rightarrow \mathbb{R}^n$ are discretized on grids characterized by grid points $\{x_i\}$ and quadrature weights $\{\omega_i\}$ for numerical integration:
\begin{equation}
    \int_{S^2} u(x) \mathrm{d}\mu(x) \approx \sum_i u(x_i) \omega_i,
\end{equation}
where $\mathrm{d}\mu(x) = \sin\vartheta \, \mathrm{d}\vartheta \, \mathrm{d}\varphi$ is the invariant measure on $S^2$.

\paragraph{Equiangular Grids}
Equiangular (lat-lon) grids use equally spaced points in spherical coordinates:
\begin{equation}
    \vartheta_i = \pi i / n_\text{lat}, \quad \varphi_j = 2\pi j / n_\text{lon},
\end{equation}
with trapezoidal quadrature weights $\omega_{ij} = (2\pi^2 / n_\text{lat} n_\text{lon}) \sin\vartheta_i$.

\paragraph{Gaussian Grids}
Gaussian grids replace the latitude grid with Gauss-Legendre nodes $\{\vartheta_i\}$ such that $\cos\vartheta_i$ are roots of the Legendre polynomial $P_{n_\text{lat}}$. This choice enables the exact integration of spherical harmonics to degree $2n_\text{lat} - 1$, making Gaussian grids particularly efficient for spectral methods.

\subsection{Spherical Harmonics}
The spherical harmonics $Y_\ell^m: S^2 \rightarrow \mathbb{C}$ form a complete orthonormal basis for $L^2(S^2)$. They are the eigenfunctions of the spherical Laplacian operator $\Delta_{S^2}$ and are defined as:
\begin{equation}
    Y_\ell^m(\vartheta, \varphi) = c_\ell^m P_\ell^m(\cos \vartheta) e^{im\varphi},
\end{equation}
where $\ell \geq 0$ is the degree, $|m| \leq \ell$ is the order, $P_\ell^m$ are the associated Legendre polynomials, and $c_\ell^m$ is a normalization constant given by:
\begin{equation}
    c_\ell^m = \sqrt{\frac{2\ell+1}{4\pi} \frac{(\ell-m)!}{(\ell+m)!}}.
\end{equation}
Any function $u \in L^2(S^2)$ can be expanded in terms of spherical harmonics via the Spherical Harmonic Transform (SHT):
\begin{equation}
    u(\vartheta, \varphi) = \sum_{\ell=0}^\infty \sum_{m=-\ell}^\ell \hat{u}_\ell^m Y_\ell^m(\vartheta, \varphi), \quad \text{where} \quad \hat{u}_\ell^m = \int_{S^2} u(x) \overline{Y_\ell^m(x)} \mathrm{d}\mu(x).
\end{equation}
The coefficients $\hat{u}_\ell^m$ form the spectral representation of the signal.

\paragraph{Implementation Details}
The Spherical Harmonics Transform can be implemented in a factorized fashion which exploits the tensor product structure of $Y_\ell^m$~\citep{Schaeffer2013}:

First, we perform a Fourier transform in longitude. For each latitude $\vartheta_k$, compute
\begin{equation}
\tilde{u}^m(\vartheta_k) = 2\pi \sum\limits_{j=0}^{n_\mathrm{lon} - 1} u(\vartheta_k, \varphi_j)\,e^{-im\varphi_j}, \qquad m = 0, \dots, M{-}1
\end{equation}
via a real-valued FFT truncated to the first $M$ modes.
Then, we perform the Legendre transform in latitude. For each order $m$, contract over the quadrature nodes:
\begin{equation}
\hat{u}_\ell^m = \sum\limits_{k=0}^{n_\mathrm{lat} - 1} \omega_k \, P_\ell^m(\cos\vartheta_k)\, \tilde{u}^m(\vartheta_k), \qquad \ell = 0, \dots, L{-}1
\end{equation}
The Legendre coefficients $P_\ell^m(\cos\vartheta_k)$ are pre-computed for a given grid and stored in a tensor of size $L{\times} M{\times} n_\mathrm{lat}$.
In order to reduce the number of operations, we further fuse the quadrature weights $\omega_j$ into this tensor.

The inverse Spherical Harmonics Transform reverses the order of operations above: first a Legendre synthesis followed by an inverse FFT. Note that no quadrature weights are involved in the inverse Legendre transformation.

\paragraph{Angular Power Spectrum}
The angular power spectrum (APS) or power spectral density (PSD) quantifies the distribution of energy across spatial scales. For a signal $u$ with spherical harmonic coefficients $\hat{u}_\ell^m$, the power at degree $\ell$ is:
\begin{equation}
    \mathrm{PSD}(\ell) = \sum_{m=-\ell}^\ell |\hat{u}_\ell^m|^2.
\end{equation}
This measures the contribution of scale $\ell$ to the total energy of the signal. The APS is rotation-invariant and provides a compact summary of the signal's spectral content.

\subsection{Vector Spherical Harmonics}
For vector fields $\mathbf{v}: S^2 \rightarrow \mathbb{R}^3$ tangent to the sphere, we employ Vector Spherical Harmonics (VSH). Any tangent vector field can be uniquely decomposed into toroidal (divergence-free) and poloidal (curl-free) components:
\begin{equation}
    \mathbf{v} = \mathbf{v}_\text{tor} + \mathbf{v}_\text{pol},
\end{equation}
where $\mathbf{v}_\text{tor} = \nabla \times (\Psi \mathbf{\hat{r}})$ and $\mathbf{v}_\text{pol} = \nabla \Phi$ for scalar potentials $\Psi, \Phi: S^2 \rightarrow \mathbb{R}$.

The VSH transform decomposes $\mathbf{v}$ into spectral coefficients $(\hat{\Psi}_\ell^m, \hat{\Phi}_\ell^m)$, enabling efficient computation of differential operators. Specifically, the \textbf{Divergence} is given by $\nabla \cdot \mathbf{v} = \Delta_{S^2} \Phi$, computed via $\widehat{\nabla \cdot \mathbf{v}}_\ell^m = -\ell(\ell+1) \hat{\Phi}_\ell^m$, and the \textbf{Curl} is $\nabla \times \mathbf{v} = \Delta_{S^2} \Psi \mathbf{\hat{r}}$, computed via $\widehat{\nabla \times \mathbf{v}}_\ell^m = -\ell(\ell+1) \hat{\Psi}_\ell^m$.
This spectral representation enables efficient computation of gradients, divergences, and curls in differential equation solvers and fluid dynamics simulations on the sphere.

\subsection{Group Convolutions}
To define a convolution operation on the sphere that generalizes the standard translation-equivariant convolution in Euclidean space, we look to group convolution. For a signal $u \in L^2(S^2)$ and a filter $k \in L^2(S^2)$, the group convolution is defined as the inner product of the signal with the rotated filter, resulting in a function defined on the rotation group $SO(3)$:
\begin{equation}
    (u \star k)(R) = \int_{S^2} u(x) k(R^{-1}x) \mathrm{d}\mu(x), \quad R \in SO(3).
\end{equation}
To obtain an output on the sphere $S^2$ rather than $SO(3)$, we restrict the rotation $R$ to the quotient space $SO(3)/SO(2) \simeq S^2$. This corresponds to fixing the rotation around the local vertical axis (usually $\gamma=0$ in Euler angles), yielding the spherical convolution:
\begin{equation}
    (u \ast k)(x) = \int_{S^2} u(x') k(R_x^{-1} x') \mathrm{d}\mu(x'),
\end{equation}
where $R_x$ is a rotation that maps the North Pole to $x$.

\subsection{The Convolution Theorem}
For zonal (isotropic) filters $k$, which depend only on the colatitude $\vartheta$ and are invariant under rotation around the $z$-axis, the spherical convolution simplifies significantly in the spectral domain. The Spherical Convolution Theorem states that the SHT of the convolution of a signal $u$ with a zonal filter $k$ is the pointwise product of their spherical harmonic coefficients:
\begin{equation}
    \widehat{(u \ast k)}_\ell^m = \sqrt{\frac{4\pi}{2\ell+1}} \hat{u}_\ell^m \hat{k}_\ell^0.
\end{equation}
This allows for efficient computation of global, isotropic convolutions by performing the operation in the spectral domain, which is the foundation of the Spherical Fourier Neural Operator (SFNO) \citep{Bonev2023}.

\subsection{Discrete-Continuous (DISCO) Convolutions}
While spectral convolutions are efficient for global isotropic filters, many applications require local, anisotropic filters. The Discrete-Continuous (DISCO) convolution \citep{Ocampo2022} addresses this by discretizing the continuous convolution integral directly. For a grid of points $\{x_j\}$ with quadrature weights $\{\omega_j\}$, the convolution is approximated as:
\begin{equation}\label{eq:disco_formal}
    (u \ast k)(x_i) \approx \sum_{j} u(x_j)\,k(R_{x_i}^{-1} x_j)\,\omega_j.
\end{equation}
Here, the filter $k$ is typically parameterized as a linear combination of local basis functions (e.g., compactly supported wavelets) on a tangent plane or disk centered at the North Pole. The term $k(R_{x_i}^{-1} x_j)$ represents the filter rotated to be centered at $x_i$ and evaluated at the source point $x_j$. This formulation allows for spatially localized and anisotropic processing while maintaining approximate rotation equivariance.

To obtain a learnable filter, $k$ is parametrized as a linear combination of basis functions $\tilde{k}_{\ell m}(x)$:
\begin{equation}\label{eq:disco_basis_expansion}
    k(x) = \sum_{\ell, m} w_{\ell m}\; \tilde{k}_{\ell m}(x).
\end{equation}
We implement multiple filter-basis functions:

\paragraph{Morlet-like wavelets}
A filter-basis inspired by Morlet-like wavelets defined on a compact disk $\vartheta' = \vartheta/\vartheta_\text{cutoff} \in [0, 1], \varphi \in [0, 2\pi)$:
\begin{equation}
    \tilde{k}_{\ell m}(\vartheta', \varphi) = h(\vartheta')\, e^{(i \pi \ell \,\vartheta' \sin \varphi)}\,e^{(i \pi m \,\vartheta' \cos \varphi)},
\end{equation}
where $h(\vartheta') = \cos^2\left(\frac{\pi}{2}\vartheta'\right)$ is the Hann windowing function. This ensures that smooth, compactly supported filters are learned while keeping the convolution tensor sparse.

\paragraph{Zernike polynomials}
A filter-basis inspired by Zernike polynomials defined on a compact disk $\vartheta' = \vartheta/\vartheta_\text{cutoff} \in [0, 1], \varphi \in [0, 2\pi)$:
\begin{equation}
    \tilde{k}_{\ell m}(\vartheta', \varphi) = Z_\ell^m(\vartheta', \varphi),
\end{equation}
where $Z_\ell^m(\vartheta', \varphi)$ are the Zernike polynomials \citep{born2013principles}. This parameterizes a filter basis that is orthogonal on the disk.

\paragraph{Piecewise linear filters}
The filter is parameterized using a tensor product of linear B-splines (hat functions) on a polar grid over the disk. Specifically, the basis functions $\tilde{k}_{\ell m}$ are products of the radial hat functions $h_{\ell}(\vartheta')$ and the angular hat functions $g_{m}(\varphi)$, centered on the nodes $(\vartheta'_\ell, \varphi_m)$. This provides a flexible, local basis that naturally handles the polar geometry of the filter kernel.

\subsubsection{Implementation Details}

Analogous to~\cite{Ocampo2022}, we define the convolution tensor
\begin{equation} \label{eq:disco_convolution_tensor}
\Psi^r_{i, (s, t)} = \omega_j\,\tilde{k}_{r}\bigl(R_{x_i}^{-1}x(\vartheta_s, \varphi_t)\bigr), 
\end{equation}
where we have flattened the kernel basis indices $\ell, m$ from \eqref{eq:disco_basis_expansion} into a single super index $r$ with total number of basis functions $K$. Note that $\Psi$ only depends on the geometry (i.e. the spherical grid and resolution). Therefore, it can be pre-computed and stored in memory as sparse tensor. 
Because the input grid is equispaced in longitude and the kernel is zonal, shifting the input field by one longitudinal grid spacing $\Delta\varphi=2\pi/\mathrm{nlon\_in}$
is equivalent to evaluating the convolution at a longitudinally shifted output point.
Using $\Psi$ can rewrite equation \eqref{eq:disco_formal} as follows:
\begin{equation} \label{eq:disco_implementation}
(u \ast k)(\vartheta_i, \varphi_j) = \sum_{r=0}^{K-1} w_r \sum\limits_{s=0}^{\mathrm{nlat\_in}-1}\sum\limits_{t=0}^{\mathrm{nlon\_in}-1}\Psi^r_{i, (s,t)}\, u\bigl(\vartheta_s, \varphi_{\bmod(t + j, \mathrm{nlon\_in})} \bigr)
\end{equation}
For multiple input and output features, we can augment the basis function weights $w_r$ accordingly similar to euclidian convolutions. 
If the output grid has a coarser resolution than the input grid (i.e. if the kernel is downsampling), the shift in $\varphi$ can be performed with stride $s=\mathrm{nlon\_in}/\mathrm{nlon\_out}$.

In all cases, the DISCO convolution kernel can be viewed as a sparse times dense matrix multiplication with an additional shift term. Because of the complicated memory access patterns, we decided to implement a custom CUDA kernel for this operation, cf.~\Cref{sec:custom_cuda_operators}.

The transpose convolution (which is also the backward of the above) can be implemented in similar manner, with summation over output indices instead of input indices in \eqref{eq:disco_implementation}.

\subsection{Spherical Attention}
Attention mechanisms can be viewed as data-dependent, non-stationary kernel smoothing. On the sphere, the continuous self-attention mechanism for a query $q$, key $k$, and value $v$ is given by \citep{Bonev2025a}:
\begin{equation}
    \mathrm{Attn}(q, k, v)(x) = \int_{S^2} \frac{\exp(q(x)^T k(x'))}{\int_{S^2} \exp(q(x)^T k(x'')) \mathrm{d}\mu(x'')} v(x') \mathrm{d}\mu(x').
\end{equation}
In \texttt{torch-harmonics}, this integral is discretized using the appropriate quadrature weights $\omega_j$:
\begin{equation}
    \mathrm{Attn}(x_i) \approx \sum_{j} \frac{\exp(q(x_i)^T k(x_j))}{\sum_{l} \exp(q(x_i)^T k(x_l)) \omega_l} v(x_j) \omega_j.
\end{equation}
Including the quadrature weights $\omega$ is crucial for accounting for the non-uniform sampling density of spherical grids (e.g., points clustering near the poles in equiangular grids), thereby ensuring that the attention mechanism approximately respects the spherical geometry and $SO(3)$ equivariance.

\paragraph{Neighborhood Attention}
To reduce computational complexity and introduce a locality inductive bias, we also implement neighborhood attention \citep{Bonev2025a}. This mechanism restricts the attention computation to a local geodesic neighborhood around each query point. It is implemented by applying a mask $M(x, x')$ to the attention scores, where $M$ acts as an indicator function: $M(x, x') = 0$ if the geodesic distance $\text{dist}(x, x') < r$ and $M(x, x') = -\infty$ otherwise. This effectively sparsifies the attention matrix while preserving local spherical symmetries.

%% file: sections/appendix/code_examples.tex
\section{Implementation Examples}
\label{app:code_examples}

This appendix provides detailed code examples demonstrating how to build spherical neural architectures using \texttt{torch-harmonics}.

\subsection{Spherical Fourier Neural Operator}
A spectral convolution layer can be implemented in a few lines:

\begin{lstlisting}[language=Python, caption=SFNO Spectral Layer]
import torch
from torch_harmonics import RealSHT, InverseRealSHT

class SpectralConv(torch.nn.Module):
    def __init__(self, nlat, nlon, num_channels):
        super().__init__()
        self.sht = RealSHT(nlat, nlon, grid="equiangular")
        self.isht = InverseRealSHT(nlat, nlon, grid="equiangular")

        # Learnable spectral weights
        self.weights = torch.nn.Parameter(
            torch.randn(size=(num_channels, num_channels, self.sht.lmax), dtype=torch.complex64)
        )

    def forward(self, x):
        # x: [batch, channels, nlat, nlon]
        coeffs = self.sht(x)  # -> spectral domain
        coeffs = torch.einsum('bclm,dcl->bdlm', coeffs, self.weights)
        return self.isht(coeffs)  # -> spatial domain
\end{lstlisting}

The forward and inverse SHT handle all coordinate transformations, while the spectral multiplication implements a global convolution.

The library already implements a Driscoll-Healy type spectral convolutions. Those are isotropic, spherically equivariant convolutions. The weights are real-valued and only depend on $m$. For adding anisotropy, \texttt{torch-harmonics} supports a generalized spectral bias term.

\begin{lstlisting}[language=Python, caption=SFNO Spectral Layer]
import torch
from torch_harmonics import SpectralConvS2

spectral_conv_layer = SpectralConvS2(
    in_shape=(181,360),
    out_shape=(180,360),
    in_channels=16,
    out_channels=32,
    grid_in="equiangular",
    grid_out="legendre-gauss",
    bias=False
)
# Example input
input_signal = torch.randn(1,16,181,360)
output = spectral_conv_layer(input_signal) #->(1,32,180,360)
\end{lstlisting}

\subsection{DISCO Convolutions for Local Processing}
DISCO convolutions enable local, anisotropic filters:

\begin{lstlisting}[language=Python, caption=DISCO Layer]
import torch
from torch_harmonics import DiscreteContinuousConvS2

conv_layer = DiscreteContinuousConvS2(
    in_channels=16,
    out_channels=32,
    in_shape=(181,360),
    out_shape=(180,360),
    kernel_shape=(3,3),
    basis_type="piecewise linear",
    grid_in="equiangular",
    grid_out="legendre-gauss",
    bias=True,
    theta_cutoff=0.2 # determines support-radius
)
input_signal = torch.randn(1,16,181,360)
output = conv_layer(input_signal) #->(1,32,180,360)
\end{lstlisting}

The layer internally manages filter rotation and quadrature, providing approximate rotation equivariance with localized receptive fields.

\subsection{Spherical Attention}
Attention on the sphere uses quadrature weights for proper integration:

\begin{lstlisting}[language=Python, caption=Spherical Attention]
import torch
from torch_harmonics import AttentionS2, NeighborhoodAttentionS2

neighborhood_attention = NeighborhoodAttentionS2(
    in_channels=256,
    out_channels=256,
    num_heads=8,
    in_shape=(181,360),
    out_shape=(180,360),
    grid_in="equiangular",
    grid_out="legendre-gauss",
    theta_cutoff=0.2,
    bias=True,
)

attention = AttentionS2(
    in_channels=256,
    out_channels=256,
    num_heads=8,
    in_shape=(181,360),
    out_shape=(180,360),
    grid_in="equiangular",
    grid_out="legendre-gauss",
    bias=False
)

k = torch.randn(1,256,181,360) #->(B,in_channels,*in_shape)
v = torch.randn(1,256,181,360) #->(B,out_channels,*in_shape)
q = torch.randn(1,256,180,360) #->(B,in_channels,*out_shape)
n_out = neighborhood_attention(q,k,v) #->(1,256,180,360)
out = attention(q,k,v) #->(1,256,180,360)
\end{lstlisting}

The attention integral employs quadrature weights to ensure the attention mechanism respects spherical geometry, accounting for varying grid cell areas.

\subsection{Hybrid Model: Combining SFNO and DISCO}
FourCastNet 3 demonstrates how to combine global and local processing:

\begin{lstlisting}[language=Python, caption=Hybrid Operator Block]
import torch
from torch_harmonics import SpectralConvS2, DiscreteContinuousConvS2

class HybridBlock(torch.nn.Module):
    def __init__(self, channels, nlat, nlon):
        super().__init__()
        self.spectral = SpectralConvS2(
            in_shape=(nlat,nlon),
            out_shape=(nlat,nlon),
            in_channels=channels,
            out_channels=channels,
            grid_in="equiangular",
            grid_out="equiangular",
            bias=True,
        )
        self.disco = DiscreteContinuousConvS2(
            in_channels=channels,
            out_channels=channels,
            in_shape=(nlat,nlon),
            out_shape=(nlat,nlon),
            kernel_shape=(3,),
            basis_type="piecewise linear",
            grid_in="equiangular",
            grid_out="equiangular",
            bias=True,
        )
        self.activation = torch.nn.GELU()

    def forward(self, x):
        # Global path
        x = x + self.activation(self.spectral(x))
        # Local path
        x = x + self.activation(self.disco(x))
        return x

input_signal = torch.randn(1,16,180,360)
hybrid_block = HybridBlock(16,180,360)
hybrid_block.forward(input_signal) #->(1,16,180,360)
\end{lstlisting}

This hybrid approach captures both large-scale wave dynamics (via SFNO) and small-scale local features (via DISCO), making it suitable for complex physical systems like atmospheric flows.

\subsection{Distributed Spectral Convolution}
The library allows for 2D domain decomposition along latitude and longitude dimensions. For this, two orthogonal processor groups need to be created. A third one is required if batch/data parallelism should also be employed. All layer-relevant data gradient collective operations are captured in the layer definitions via custom autograd mechanics. However, since weight gradients are handled separately by PyTorch, additional reductions have to be registered. The example below shows how this can be achieved with \texttt{torch.distributed} and \texttt{torch-harmonics} layers such as SpectralConvS2. The following example is implemented for GPUs, and we assume that the environment variables \texttt{RANK}, \texttt{LOCAL\_RANK}, \texttt{WORLD\_SIZE}, \texttt{MASTER\_ADDR} and \texttt{PORT} have been set according to the PyTorch distributed computing documentation.

\begin{lstlisting}[language=Python, caption=Distributed spectral convolution]
import os

import torch
import torch.distributed as dist
from torch.distributed.device_mesh import init_device_mesh
from torch.nn.parallel import DistributedDataParallel as DDP

import torch_harmonics.distributed as thd
from torch_harmonics.distributed import DistributedSpectralConvS2

# initialize the world process group
world_rank = int(os.environ.get("RANK"))
world_size = int(os.environ.get("WORLD_SIZE"))
dist.init_process_group(
        backend="nccl",
        init_method=None,
        rank=world_rank,
        world_size=world_size)
local_rank = int(os.environ.get("LOCAL_RANK"))

# better set a device:
device = torch.device(f"cuda:{local_rank}")
torch.cuda.set_device(device.index)

# initialize device mesh
data_dim_size = 8 # number of GPUs in data direction
polar_dim_size = 2 # number of GPUs in polar / latitude direction
azimuth_dim_size = 4 # number of GPUs in azimuth / longitude direction
# sanity checks
assert world_size == data_dim_size * polar_dim_size * azimuth_dim_size
mesh = init_device_mesh("cuda", [data_dim_size, polar_dim_size, azimuth_dim_size], mesh_dim_names=["data", "lat", "lon"])

# now we can initialize the azimuth and polar comm groups for torch harmonics: this will add the respective comm groups created by the mesh into a hash lookup table which is used by TH to find the corresponding ones:
thd.init(mesh.get_group("lat"), mesh.get_group("lon"))

# Now we can define a distributed SpectralConvS2 layer:
# note that the in and out shapes should be the global shapes, not the decomposed ones!
distributed_spectral_conv = DistributedSpectralConvS2(
    in_shape=(360, 720),
    out_shape=(360, 720),
    in_channels=256,
    out_channels=256,
    grid_in="equiangular",
    grid_out="equiangular",
    bias=True,
).to(device)

# now initialize DDP for the batch reduction
# make sure to only use the data group here, not the world group
model_ddp = DDP(
        distributed_spectral_conv,
        device_ids=[device],
        output_device=device,
        process_group=mesh.get_group("data"),
)

# now we need to ensure that the weight gradients are reduced properly: to do so, we need to understand how they are shared between ranks: for the spectral convolution, the bias is fully decomposed along lon and lat and so the gradients for the bias should not be reduced along those dimensions (DDP takes care of the data group reductions). Therefore, we do not need to do anything for the bias. However, the weight is only decomposed in lat-direction and thus shared in lon direction. Therefore, we need to reduce the gradient of this along that direction. The easiest way to do this is to register a post accumulation gradient hook like this:
def _longitude_reduction_hook(param: torch.Tensor):
    if param.grad is not None:
        dist.all_reduce(
            param.grad,
            group=mesh.get_group("lat"),
            op=dist.ReduceOp.SUM
        )
    return

# register the hook to fire automatically after a gradient is computed and accumulated
model_ddp.distributed_spectral_conv.weight.register_post_accumulate_grad_hook(_longitude_reduction_hook)

# now we assume we already have a global input tensor (for example loaded from a corresponding dataset). We need to split it across ranks. This can be done by using split_tensor_along_dim
inp = torch.randn(1,256,360,720,
    dtype=torch.float32,device=device)

# assume inp has shape B, C, NLAT, NLON:
# split in lat direction
inp_split_list = thd.split_tensor_along_dim(inp, dim=-2, num_chunks=thd.polar_group_size()) # one can also use mesh.get_group("lat").size() here
# take only the data belonging to the corresponding polar rank:
inp_split = inp_split_list[thd.polar_group_rank()]
# split in lon direction
inp_split_list = thd.split_tensor_along_dim(inp_split, dim=-1, num_chunks=thd.azimuth_group_size())
inp_split = inp_split_list[thd.azimuth_group_rank()]

# now we can feed this tensor into our distributed layer
out_split = model_ddp(inp_split)

# we can use this output compute losses, backward passes and optimizer updates and each rank will receive correct gradients.
\end{lstlisting}

\subsection{Shallow Water Equations Solver}
\label{app:swe_solver}

The shallow water equations govern the evolution of a thin fluid layer on a rotating sphere and serve as a simplified model for atmospheric dynamics:
\begin{align}
    \frac{\partial \mathbf{v}}{\partial t} &= -(\zeta + f)\mathbf{k} \times \mathbf{v} - \nabla\left(gh + \frac{|\mathbf{v}|^2}{2}\right), \label{eq:swe_momentum} \\
    \frac{\partial h}{\partial t} &= -\nabla \cdot (h \mathbf{v}), \label{eq:swe_continuity}
\end{align}
where $\mathbf{v}$ is the velocity field, $h$ is the fluid height, $\zeta = \nabla \times \mathbf{v}$ is the relative vorticity, $f = 2\Omega \sin\vartheta$ is the Coriolis parameter, and $g$ is gravitational acceleration.

A complete implementation of a differentiable shallow water equations solver:

\begin{lstlisting}[language=Python, caption=Shallow Water Solver using torch-harmonics]
import torch
from torch_harmonics.quadrature import clenshaw_curtiss_weights
from torch_harmonics.sht import RealSHT, InverseRealSHT, RealVectorSHT, InverseRealVectorSHT

class ShallowWaterSolver(torch.nn.Module):
    def __init__(self, nlat, nlon, dt, lmax=None, mmax=None,
                 radius=6.37122e6, omega=7.292e-5,
                 gravity=9.80616, havg=1e4, hamp=120.0):
        super().__init__()

        self.dt = dt
        self.radius, self.gravity = radius, gravity
        self.havg, self.hamp = havg, hamp

        self.sht  = RealSHT(nlat, nlon,
            lmax=lmax, mmax=mmax, grid="equiangular")
        self.isht = InverseRealSHT(nlat, nlon,
            lmax=lmax, mmax=mmax, grid="equiangular")
        self.vsht = RealVectorSHT(nlat, nlon,
            lmax=lmax, mmax=mmax, grid="equiangular")
        self.ivsht = InverseRealVectorSHT(nlat, nlon,
            lmax=lmax, mmax=mmax, grid="equiangular")

        lmax, mmax = self.sht.lmax, self.sht.mmax
        cost, _ = clenshaw_curtiss_weights(nlat, -1, 1)
        lats = -torch.arcsin(cost)

        l = torch.arange(0, lmax,dtype=torch.float64)
        l = l.reshape(lmax,1).expand(lmax, mmax)
        self.lap = -l * (l + 1) / radius**2
        self.invlap = torch.where(l > 0, -radius**2 /
            (l * (l + 1)), torch.zeros_like(l))
        self.f = 2 * omega * torch.sin(lats).reshape(nlat, 1)
        self.hyperdiff = torch.exp((-dt / 2 / 3600.) * (self.lap / self.lap[-1, 0])**4)

    def vrtdivspec(self, uv_grid):
        return self.lap * self.radius * self.vsht(uv_grid)

    def getuv(self, vrtdiv_spec):
        return self.ivsht(self.invlap * vrtdiv_spec / self.radius)

    def rhs(self, uspec):
        dudt    = torch.zeros_like(uspec)
        phi     = self.isht(uspec[0])
        uv      = self.getuv(uspec[1:])
        abs_vrt = self.isht(uspec[1]) + self.f

        fs         = self.vrtdivspec(uv * abs_vrt)
        dudt[1]    = -fs[1]
        dudt[2]    =  fs[0]
        dudt[0]    = -self.vrtdivspec(uv * phi)[1]
        dudt[2]   -= self.lap * \
            self.sht(phi + 0.5 * (uv[0]**2 + uv[1]**2))

        return dudt

    def timestep(self, uspec, nsteps):
        history = torch.zeros(3, *uspec.shape, dtype=uspec.dtype)
        new, now, old = 0, 1, 2

        for i in range(nsteps):
            history[new] = self.rhs(uspec)
            if i == 0:
                history[now] = history[old] = history[new]
            elif i == 1:
                history[old] = history[new]

            uspec = uspec + self.dt * \
                ((23./12.) * history[new] - (16./12.) * \
                history[now] + (5./12.) * history[old])
            uspec[1:] = self.hyperdiff * uspec[1:]

            new = (new - 1) % 3
            now = (now - 1) % 3
            old = (old - 1) % 3

        return uspec

    def initial_condition(self, mach=0.1):
        uspec = torch.randn(3, self.sht.lmax, self.sht.mmax,
            dtype=torch.complex128)
        uspec[0]      *= self.gravity * self.hamp / self.sht.lmax
        uspec[0, 0, 0] = (4 * torch.pi)**0.5 * self.havg * \
            self.gravity
        uspec[1:]     *= (mach *
            (self.gravity * self.havg)**0.5 /
            self.radius / self.sht.lmax)
        return torch.tril(uspec)

nlat, nlon = 64, 128
solver = ShallowWaterSolver(nlat, nlon, dt=400.)
uspec  = solver.initial_condition()
uspec  = solver.timestep(uspec, nsteps=1296)
uv     = solver.getuv(uspec[1:])
speed  = torch.sqrt(uv[0]**2 + uv[1]**2)
\end{lstlisting}

This solver leverages the VSHT to compute divergence and curl efficiently in spectral space, avoiding numerical instabilities common in finite-difference schemes.

%% file: sections/appendix/performance_optimization.tex
\subsection{Performance Optimizations for Operators in PyTorch}

In this section we briefly describe what performance optimizations we have applied to some of the \texttt{torch-harmonics} kernels implemented in pure PyTorch.

\subsubsection{Memory Layout Optimizations for SHT}

The computational bottleneck of the SHT is the Legendre transform, which reduces to a batched matrix-matrix multiplication. For the forward transform, the contraction is over the latitudinal index $k$; for the inverse, it is over the degree index $\ell$. To maximize arithmetic intensity and memory throughput on GPU architectures, the implementation applies two key layout choices:

Stride-1 contraction index. Before each Legendre contraction, the two trailing tensor dimensions are transposed so that the summation index ($n_\mathrm{lat}$ in the forward transform, $L$ in the inverse) resides in the fastest-varying (stride-1) memory position. Concretely, in the forward case the intermediate Fourier coefficients are stored as $(\ldots, m, k)$ with $k$ stride-1, and in the inverse case the spectral coefficients are stored as $(\ldots, m, \ell)$ with $\ell$ stride-1.

The precomputed Legendre weight tensors are stored in a compatible layout so that the contraction index is stride-1 in both operands. In the forward transform, the quadrature-weighted associated Legendre polynomials $W_{m,\ell,k} = \omega_k\,P_\ell^m(\cos\vartheta_k)$ are stored with $k$ as the fastest index. In the inverse transform, the synthesis tensor $P_\ell^m(\cos\vartheta_k)$ is stored as $(m, k, \ell)$ with $\ell$ being the fastest index. This ensures that the inner loop of the batched contraction reads both input and kernel from contiguous memory.

Together, these layout choices allow the Legendre step to be cast as a high-performance batched GEMM, fully exploiting the memory hierarchy and tensor core capabilities of modern GPUs.

\subsection{Implementation Details for Distributed Operations}

\subsubsection{Distributed SHT}
The distributed implementation partitions the computation across a two-dimensional process grid of size $p_\mathrm{lat} \times p_\mathrm{lon} $, where $p_\mathrm{lat}$ processes decompose the latitudinal (polar or $\vartheta$) dimension and $p_\mathrm{lon}$ processes decompose the longitudinal (azimuthal or $\varphi$) dimension. Communication within each group is performed via NCCL all-to-all collectives. This is analogous to pencil decompositions in multi-dimensional Fourier transformations.

In the initial data layout, each process owns a local tile of the spatial grid of size $n_\mathrm{lat} / p_\mathrm{lat} \times n_\mathrm{lon} / p_\mathrm{lon}$, together with all $C$ channels. We denote a distributed dimension by underlining it: the initial layout is $(C, \underline{n_\mathrm{lat}}, \underline{n_\mathrm{lon}})$.

\paragraph{Forward Transform}
The forward distributed SHT proceeds through the following sequence of transpositions and local computations:

Step 1 — Azimuthal all-to-all (making $\mathrm{lat}$ local): Starting from the layout  $(C, \underline{n_\mathrm{lat}}, \underline{n_\mathrm{lon}})$, an all-to-all transposition over the azimuthal process group redistributes the longitudinal dimension into each rank while distributing the channel dimension across ranks:

\begin{equation}
(C, \underline{n_\mathrm{lat}}, \underline{n_\mathrm{lon}}) \xrightarrow{\text{all-to-all}-\varphi} (\underline{C}, \underline{n_\mathrm{lat}, n_\mathrm{lon}})
\end{equation}

Each process now holds the full longitudinal extent for its local latitude slab and a subset of the channels.

Step 2 — Local FFT. Each process independently applies a real-to-complex FFT along the longitudinal axis and truncates to $M$ modes:

\begin{equation}
(\underline{C}, \underline{n_\mathrm{lat}}, n_\mathrm{lon}) \xrightarrow{\mathcal{F}-\varphi} (\underline{C}, \underline{n_\mathrm{lat}}, M)
\end{equation}

Step 3 — Azimuthal all-to-all (distribute $m$, restore $C$): 
a second all-to-all over the azimuthal group distributes the spectral order $m$ and restores the channel dimension:

\begin{equation}
(\underline{C}, \underline{n_\mathrm{lat}}, M) \xrightarrow{\text{all-to-all}-\varphi} (C, \underline{n_\mathrm{lat}}, \underline{M})
\end{equation}

Step 4 — Polar all-to-all (make $n_\mathrm{lat}$ local): 
an all-to-all over the polar process group gathers the full latitudinal extent at the cost of distributing the channel dimension:

\begin{equation}
(C, \underline{n_\mathrm{lat}}, \underline{M}) \xrightarrow{\text{all-to-all}-\theta} (\underline{C}, n_\mathrm{lat}, \underline{M})
\end{equation}

Step 5 — Local Legendre transform: 
with the full latitudinal range available, each process performs the weighted Legendre projection locally, using the stride-1 memory layout optimizations described above. The precomputed weight tensor $W_{m,\ell,k} = \omega_k \, P_\ell^m(\cos\vartheta_k)$ is stored only for the local shard of $m$:

\begin{equation}
\hat{u}_\ell^m = \sum\limits_{k=0}^{n_\mathrm{lat} - 1} W_{m,\ell,k} \, \tilde{u}^m(\vartheta_k), \qquad (\underline{C}, n_\mathrm{lat}, \underline{M}) \xrightarrow{\mathcal{L}_\theta} (\underline{C}, L, \underline{M})
\end{equation}

Step 6 — Polar all-to-all (distribute $\ell$, restore $C$): 
a final all-to-all over the polar group distributes the degree $\ell$ and restores the full channel dimension:

\begin{equation}
(\underline{C}, L, \underline{M}) \xrightarrow{\text{all-to-all}-\theta} (C, \underline{L}, \underline{M})
\end{equation}

The output tensor of shape $(C, \underline{L}, \underline{M})$ contains the spectral coefficients with both $\ell$ and $m$ distributed.

Note that the Legendre transformation can in principle be performed with a distributed matrix multiplication. However, in this case, the results will deviate from the corresponding serial operation for the same input tensors because of order of operation differences. The all-to-all approach mitigates this problem and outputs are bit-wise identical to the corresponding Serial Harmonics Transform. 

\paragraph{Inverse Transform}
In order to compute the inverse transform we simply reverse the above sequence. Note that for the inverse transform, no quadrature weights are needed in the Legendre transformation.

\paragraph{Forward and Inverse Vector SHT}
The vector SHT can be decomposed into scalar SHT and linear combinations of vector as well as real and imaginary components the transformed fields. Therefore, the same strategy described above applies to this case as well.

\subsubsection{Distributed DISCO Convolution}
The input data is distributed on the same 
$p_\mathrm{lat}\times p_\mathrm{lon}$ process grid used for the SHT, starting in the layout (we ignore batch sizes and potential other indices before the channel dim since the procedure vectorizes over those).
$(C, n_\mathrm{lat}, n_\mathrm{lon})$

\paragraph{Splitting the convolution tensor.} The shift in the longitudinal index in equation~\eqref{eq:disco_implementation} couples all longitudes, so the full longitudinal extent must be locally available on each process. The latitudinal dimension, however, can be decomposed: the precomputed sparse tensor 
$\Psi$ (cf. \eqref{eq:disco_convolution_tensor})
can be split along the input latitude index so that each polar rank holds only the non-zero entries whose input latitude falls into its local slab $s(p) \in [s_\mathrm{start}(p), s_\mathrm{end}(p)[$
The column indices are rewritten to refer to the local input tile, yielding a per-rank sparse tensor $\Psi^r_{i,(s(p),t)}$
Note that it still addresses all output latitudes $\theta_i$, since the kernel support can extend across slab boundaries.

\paragraph{Forward convolution.} The distributed forward pass proceeds as follows:

Step 1 --- Azimuthal all-to-all (make 
$\varphi$ local): 
perform all-to-all over the azimuthal process group in order to gather the full longitudinal extent on each rank, distributing the channel dimension in return: 
\begin{equation} (C, \underline{\mathrm{nlat\_in}}, \underline{\mathrm{nlon\_in}}) \xrightarrow{\text{all-to-all}-\varphi} (\underline{C}, \underline{\mathrm{nlat\_in}}, \mathrm{nlon\_in}).
\end{equation}

Step 2 --- Apply local DISCO kernel: 
each polar rank $p$ applies its local sparse tensor $\Psi^r_{i,(s(p),t)}$ to its input slab using the custom CUDA kernel described in Section~\ref{sec:custom_cuda_operators}. Because 
the tensor addresses all output latitudes, this produces a partial output of full latitudinal extent $\mathrm{nlat\_out}$., containing only the contributions from the local input rows: 
\begin{equation} 
u^{(p)}_{r,i,j} = \sum\limits_{s=s_\mathrm{start}^{(p)}}^{s_\mathrm{end}^{(p)}-1} \sum\limits_{t=0}^{\mathrm{nlon\_in}-1} \Psi^r_{i,(s(p),t)}\, u\bigl(\vartheta_s, \varphi_{\bmod(t+j, \mathrm{nlon\_in})}\bigr). 
\end{equation}

Step 3 --- All-reduce and scatter over the polar group. 
The partial outputs from all polar ranks are summed via an all-reduce over the polar process group, recovering the complete contraction 
The result is then scattered along the output latitude dimension so that each rank holds only its local output slab: 
\begin{equation} 
(\underline{C}, K, \mathrm{nlat\_out}, \mathrm{nlon\_out}) \xrightarrow{\text{all-reduce}-\vartheta + \text{scatter}-\vartheta} (\underline{C}, K, \underline{\mathrm{nlat\_out}}, \mathrm{nlon\_out}), 
\end{equation}
where K denotes the number of kernel basis functions.

Step 4 --- Azimuthal all-to-all:
a second all-to-all over the azimuthal group restores the channel dimension and re-distributes the longitudinal dimension: \begin{equation} 
(\underline{C}, R, \underline{\mathrm{nlat\_out}}, \mathrm{nlon\_out}) \xrightarrow{\text{all-to-all}-\varphi} (C, R, \underline{\mathrm{nlat\_out}}, \underline{\mathrm{nlon\_out}}).
\end{equation}

Step 5 --- Channel mixing: 
the basis dimension r is contracted with the learned weights 
$w_{r,\mathrm{cin}, \mathrm{cout}}$ (augmented for multiple input/output features) to produce the final output in the original distributed layout.
This step is entirely local and requires no communication.

\paragraph{Transpose convolution.} 
For the transpose convolution, the order is reversed. The channel mixing (with transposed weights) is applied first. After an azimuthal all-to-all to make $\varphi$ local, the full input latitude extent is obtained via an all-gather over the polar group. Each rank then applies the transposed tensor $\Psi^{r\,T}_{i,(s(p),t)}$ --- and sums over output indices rather than input indices as in equation~\eqref{eq:disco_implementation} --- to produce its local output latitude slab directly, without requiring a subsequent reduction. A final azimuthal all-to-all restores the original distributed layout.

\subsection{Custom CUDA Operators}\label{sec:custom_cuda_operators}

One of the development goals is to keep \texttt{torch-harmonics} as high-level as possible, enabling users to implement and test their own ideas quickly. Therefore, many kernels in the library are implemented in PyTorch, which directly translate into efficient CUDA or CPU kernels. For some kernels however, most notably DISCO and spherical neighborhood attention, PyTorch does not provide the required tools to implement those efficiently at high-level. For those cases, PyTorch provides a way to map custom kernels written in CUDA or C++ into the PyTorch namespace, including the registration of  corresponding backward kernels.

We implemented custom CUDA kernels for the forward and backward passes of the DISCO and spherical neighborhood attention transformations. Fundamentally, both transformations can be
modeled as sparse graph aggregations with appropriate quadrature weighting.

We define the input data as a set of $N$ four-dimensional tensors\footnote{$N=1$ for disco and $N=3$ (query, key and value), for attention} and we denote
their dimensions as $B \times C \times H \times W$
(Batch, Channel, Height, Width). Let the set of input tensors be:

$$ X^{(n)} \in \mathbb{R}^{B \times C \times H_{in} \times
W_{in}} \mid n = 0 \dots N-1 $$

For each output spatial site $(h_o, w_o)$, the operation gathers $C$-dimensional
feature vectors from a set of source sites across the inputs tensors and reduces
them to form the output vector. The connectivity between output sites and source
sites is represented as a sparse adjacency matrix stored in compressed sparse row
(CSR) format.
Conceptually, it encodes the edges of a sparse graph that links each output site
to the set of input sites $(h_i, w_i)$ from which it aggregates features.

The computation of the output tensor $Y \in \mathbb{R}^{B \times C \times
H_{out} \times W_{out}}$ can be formalized as:

$$ Y_{b, :, h_o, w_o} = \text{Reduce}\left( \{ X^{(n)}_{b, :, h_i, w_i} \mid
(h_i, w_i) \in \text{CSR}(h_o, w_o), \, n = 0 \dots N-1 \} \right) $$

The number of input tensors and the specific aggregation rules differ between
DISCO and Attention, and they also vary between forward and backward passes.
While the underlying computational structure is sufficiently similar to
motivate and explain the same family of kernel level optimizations, we follow a slightly different approach for DISCO and Attention layers.

\subsubsection{Spherical Neighborhood Attention}
The performance is dominated by irregular memory gathers of feature vectors.
Consequently, performance is bound by global memory bandwidth. To maximize
throughput, we enforce memory coalescence by ensuring that the dense dimension
(channels) is stored contiguously. This requires to process the tensors in a
channel-last (BHWC) layout.

However, since the surrounding code infrastructure relies on the standard
channel-first (BCHW) layout, we permute tensors immediately before and
after kernel execution:

\begin{enumerate}
\item Permute inputs from from BCHW to BHWC.
\item Launch the Attention kernel.
\item Permute the output from BHWC to BCHW.
\end{enumerate}

As these permutations occur at every invocation, their efficiency is critical
for the overall transformations performance. We compared with the
standard PyTorch \verb|permute()| operator. However, we found it to be a
significant bottleneck, achieving only $0.5$ TB/s on a GB200 GPU for $\sim$2.0
GB of data. This is not surprising, as the operator is a general-purpose
implementation, designed for maximum flexibility across a wide variety of cases.
By implementing custom CUDA kernels specialized for the specific BCHW
$\leftrightarrow$ BHWC transpose, we achieved $\sim$6.2 TB/s on the same
hardware, an order-of-magnitude improvement that renders the permutation cost
negligible relative to the aggregation kernels.

\paragraph{Kernel Architecture} Our aggregation kernels parallelize over spatial
sites. Each output site is independently processed by a group of threads that
reads the appropriate vectors ($x^{n}_{b,h_i,w_i,:}$) from the input tensors,
performs the reduction operation and then writes the resulting vector
($y_{b,h_o,w_o,:}$) to the output tensor.  Threads within the group are mapped
consecutively along the channel index to ensure fully coalesced loads and
stores.

To minimize redundant global-memory accesses and improve instruction-level
parallelism, we maintain the output vector (reduction accumulator) and
frequently accessed vectors in thread-local registers. However, given the hard
constraint on register file size, there is an upper bound on the channel
dimension that permits full register residency.  To handle arbitrary channel
widths (C), we implemented two kernel variants:

\begin{enumerate}

\item Specialized variant (register-resident): optimized for cases where the
feature vectors fit entirely within the registers of a thread group. This
variant utilizes variable thread group sizes (from 4 up to a full block of 1024
threads) to match the channel dimension.

\item General variant (shared memory): A fallback for large C, where vectors are
staged in shared memory.  This variant uses a fixed warp size to maximize
occupancy.

\end{enumerate}

Moreover, where data alignment constraints are met, the kernels leverage
vectorized 128-bit load and store instructions (e.g., \verb|float4|) to improve
memory access efficiency.
Both variants are implemented as templated functions and are instantiated for
both 32- and 128-bit accesses and for  dimensions up to 
C=$16384$,
covering the vast majority of practical workloads with
the highly optimized path. At runtime, the most suitable instance is selected
and dispatched based on the actual properties of the input data.

\paragraph{Load Balancing} The CSR structure often represents spatial
neighborhoods with a highly irregular distribution of neighbor counts. We
observed variations spanning up to three orders of magnitude, which leads to
substantial load imbalance: thread groups assigned to dense neighborhoods may be
scheduled late and create a tail effect that keeps the GPU underutilized during
the final phase of execution.

To mitigate this, we presort the CSR rows in descending order of length before
running the kernels. Thread groups then process spatial sites according to this
ordering, ensuring that the most computationally expensive sites are processed
first. This provides more opportunity for overlapping their longer computations
with groups processing smaller neighborhoods, resulting in improved GPU
utilization. In our experiments, this strategy yielded up to $20\%$ runtime
improvement on a GB200. Since the reordering depends only on the CSR row length
($H_{out}$ elements), its cost is negligible.

\subsubsection{DISCO convolution}

As in the Attention case, the performance is dominated by the gathering and
reduction of feature vectors. However, there is an important difference between
the two transformations. In Attention, the reductions require combining some
feature vectors with the dot products of others. To compute these dot products
efficiently, the tensors are stored in a channel last layout and CUDA threads
are mapped along the channel dimension.

In contrast, DISCO performs reductions across vectors independently for each
channel and does not require operations that combine all elements of a feature
vector, such as dot products. Instead, the reduction is applied element wise
across the input vectors. This structure allows the computation to be
parallelized directly across the elements of the feature vectors.

For this reason, in DISCO the input and output tensors are processed in their
native BCHW layout, without performing permutations around the kernel. In this
case, we adopt a thread to data mapping that is orthogonal to that used for
Attention, mapping threads along the rows of the tensors rather than along the
channel dimension.  As described below, this organization allows to limit
redundant reads of feature vectors when scanning the list of vectors that must
be reduced and, in the backward pass, to significantly reduce the number of
atomic operations required to scatter gradient contributions.

In the forward pass, channels from different vectors of the input tensor are
gathered and reduced into the corresponding channel of a single vector of the
output tensor. The set of input vectors depends only on the latitude of the
output vector. Moreover, many consecutive vectors may lie within the same input
latitude row. Therefore, each CTA is mapped to an output row, which is
accumulated in registers while scanning the input rows and written to the
output tensor once processing is complete. The input latitude rows are stored
in shared memory and reused as long as the index of the current input vector
refers to that row. In this way, each input feature vector is read from global
memory only once for each latitude of the output tensor that requires it.

The backward transformation is structurally dual. In this case, one CTA owns a
single input gradient row and scatters contributions to the output gradient.
The input row is read once and kept in registers. As long as the scatter
operations involve the same output row, the accumulation is performed in shared
memory. To avoid costly atomic operations within the CTA, the shared buffer is
allocated larger than the size of the output row in order to convert potential
intra-CTA scatter collisions into non-overlapping writes inside separate chunks
of the buffer. Analogously to the forward case, when a new output row must be
processed, the row currently stored in shared memory is flushed to the output
gradient. When flushing, the oversized buffer is folded back to recover the
correct output row.  Since different CTAs corresponding to distinct input rows
may update the same output row, atomic operations are required for these global
writes.